\documentclass[letterpaper, conference]{ieeeconf}  

\IEEEoverridecommandlockouts                              

\usepackage{graphics} 
\usepackage{epsfig} 
\usepackage{amsmath} 
\usepackage{amsfonts}
\usepackage{amssymb}  
\usepackage{cleveref}
\usepackage[table]{xcolor}
\usepackage{mathtools}
\usepackage{booktabs}
\usepackage{flushend}
\usepackage{siunitx}
\newcommand{\defeq}{\vcentcolon=}

\usepackage{bbm}
\usepackage{comment}
\usepackage{subcaption}
\usepackage[font={footnotesize}]{caption}

\crefname{figure}{Fig.}{Fig.}
\Crefname{figure}{Figure}{Figures}
\usepackage[utf8]{inputenc}
\usepackage{pgfplots}
\usepackage{pgfplotstable}

\pgfplotstableset{
    color cells/.style={
        col sep=comma,
        string type,
        postproc cell content/.code={%
                \pgfkeysalso{@cell content=\rule{0cm}{2.4ex}\cellcolor{black!##1}\pgfmathtruncatemacro\number{##1}\ifnum\number>50\color{white}\fi##1}%
                },
        
    }
}

\pgfplotstableset{
    color cells/.style={
        col sep=comma,
        string type,
        postproc cell content/.code={%
                \pgfkeysalso{@cell content=\rule{0cm}{2.4ex}%
                \pgfmathsetmacro\y{min(100,max(0,abs(round(##1 * 0.5))))}%
                \ifnum##1<0\edef\temp{\noexpand\cellcolor{blue!\y}}\temp\fi%
                \ifnum##1>0\edef\temp{\noexpand\cellcolor{red!\y}}\temp\fi%
                \pgfmathtruncatemacro\x\y%
                \ifnum\x>50 \color{white}\fi%
                ##1}%
                }
        columns/x/.style={
            column name={},
            postproc cell content/.code={}
        }
    }
}
\usepackage{collcell}
\usepackage{array}
\usepackage{tikz}
\usepackage{pgfkeys}
\usepackage{graphicx}
\pgfplotsset{compat=newest}
\usepgfplotslibrary{groupplots}
\usepgfplotslibrary{dateplot}

\title{\LARGE \bf
A Composable Framework for Policy Design, Learning, and Transfer Toward Safe and Efficient Industrial Insertion
}

\author{Rui Chen, Chenxi Wang, Tianhao Wei, Changliu Liu
\thanks{$^{1}$ R. Chen, C. Wang, T. Wei, and C. Liu are with Carnegie Mellon University, Pittsburgh, PA, USA.  
        {\tt\small ruic3, chenxiwa, twei2, cliu6@andrew.cmu.edu},}%
}

\begin{document}

\maketitle
\thispagestyle{empty}
\pagestyle{empty}

\begin{abstract}

Delicate industrial insertion tasks (e.g., PC board assembly) remain challenging for industrial robots. The challenges include low error tolerance, delicacy of the components, and large task variations with respect to the components to be inserted. To deliver a feasible robotic solution for these insertion tasks, we also need to account for hardware limits of existing robotic systems and minimize the integration effort. This paper proposes a composable framework for efficient integration of a safe insertion policy on existing robotic platforms to accomplish these insertion tasks. The policy has an interpretable modularized design and can be learned efficiently on hardware and transferred to new tasks easily. In particular, the policy includes a safe insertion agent as a baseline policy for insertion, an optimal configurable Cartesian tracker as an interface to robot hardware, a probabilistic inference module to handle component variety and insertion errors, and a safe learning module to optimize the parameters in the aforementioned modules to achieve the best performance on designated hardware. The experiment results on a UR10 robot show that the proposed framework achieves safety (for the delicacy of components), accuracy (for low tolerance), robustness (against perception error and component defection), adaptability and transferability (for task variations), as well as task efficiency during execution plus data and time efficiency during learning.

\end{abstract}


\section{Introduction}

In modern industrial assembly, robotic arms are widely used in many tasks. 
However, it remains challenging for robots to be applied to delicate multiple-contact insertion where the components can be easily bent or damaged. Specifically, this paper considers the task to insert integrated circuit (IC) chips to printed circuit (PC) boards where the IC chips have multiple delicate pins and have low error tolerance during insertion. Our goal is to develop a robotic insertion framework that (a) ensures safety by preventing physical damage to either the chip or the board, (b) achieves high performance in terms of accuracy and efficiency, (c) handles perception errors and component defection, (d) is able to transfer to different tasks and hardware, and (e) is easy to interpret and maintain. 
To meet these goals, we choose to design the framework to be composable, which consists of modules that can be selected and assembled in various combinations. 

Challenges to design such a framework arise from both task specifications and hardware limitations. Regarding the insertion task, the pins on the IC chips to be assembled can be easily bent upon collision with the PC board. 
Since the defects can hardly be detected, it is difficult for the robot to determine whether the defective chips are still valid for insertion and if so, how to adjust the insertion routine. Another challenge is that insertion tasks can vary widely in terms of component size and insertion tolerance. It is non-trivial to transfer the framework to various tasks without tedious tuning for each task instance. Besides task-related obstacles, the hardware implementation is also challenging. 
First, considering the detection of the workspace, the perception system might not be accurate enough to localize the PCB holes within the error tolerance for insertion. Second, there are communication delays in different modules in the robotic system, which imposes additional difficulty on real-time hardware control.

\begin{figure}[t]
\centering
\begin{subfigure}[b]{\linewidth}
     \centering
     \includegraphics[width=\linewidth]{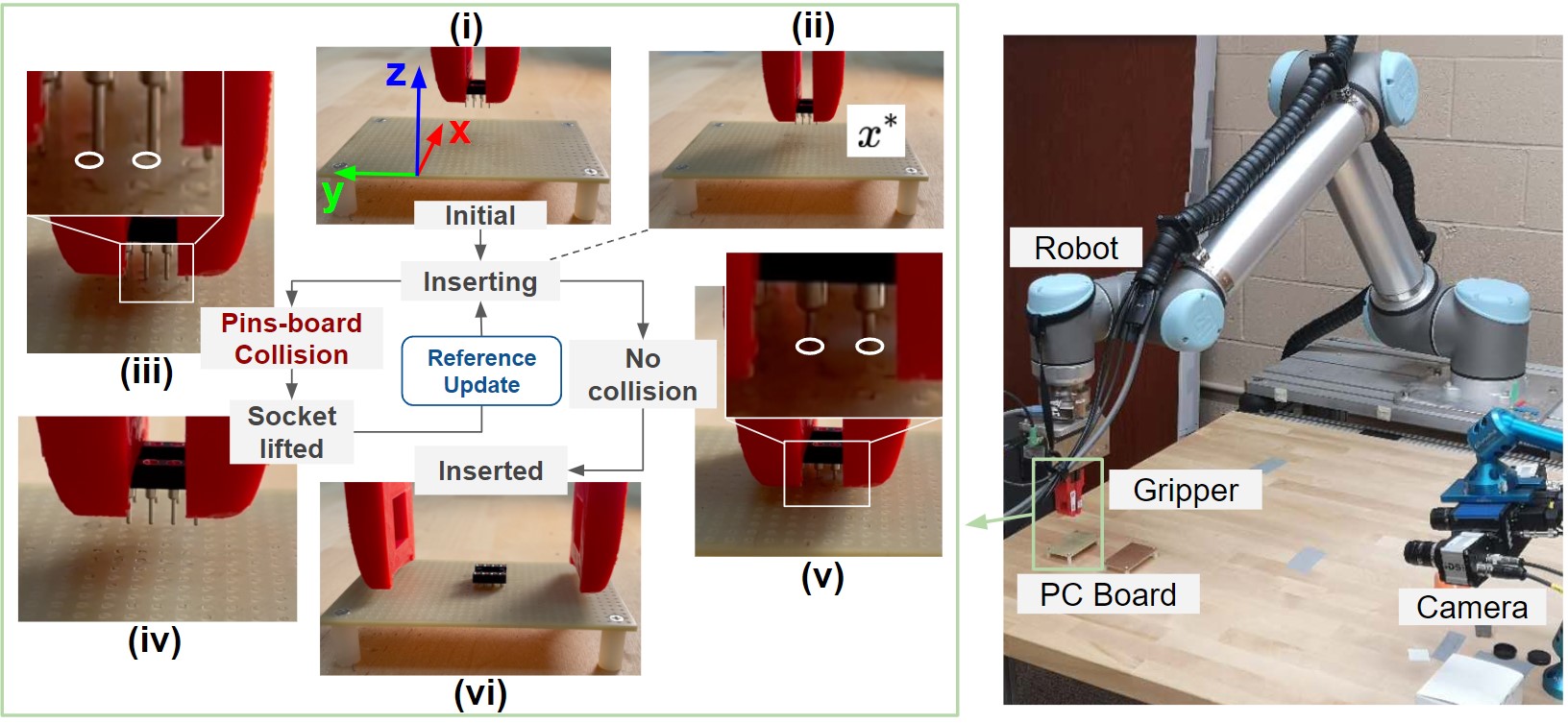}
     \caption{Industial insertion task.}
     \label{fig:task_overall}
\end{subfigure}
\vfill
\vspace{5pt}
\begin{subfigure}[b]{\linewidth}
     \centering
     \includegraphics[width=\linewidth]{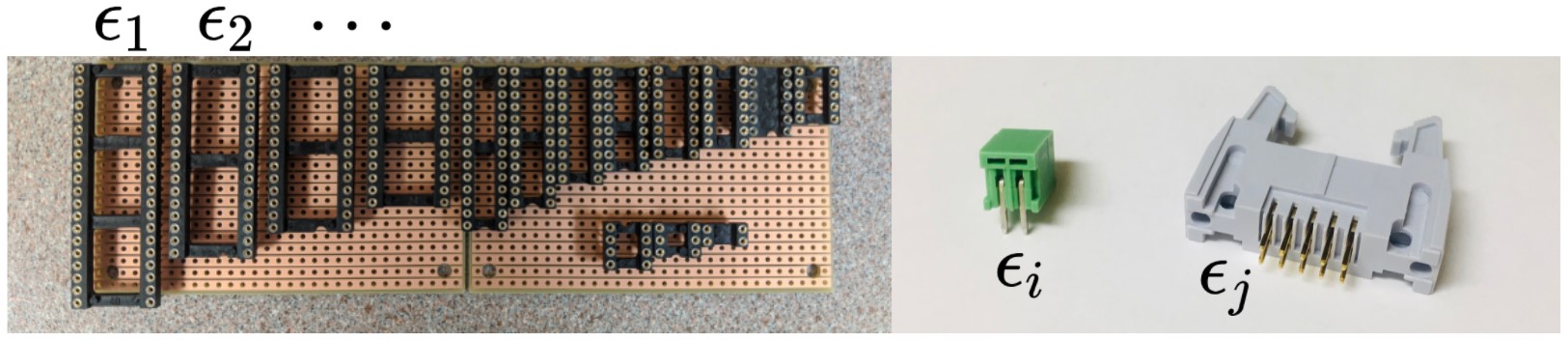}
     \caption{Variation of components.}
     \label{fig:task_variance}
\end{subfigure}
\caption{Illustration of (a) a delicate industrial insertion task and (b) variation of components to insert. (a) (right) shows the system setup. (a) (left) shows the insertion procedure. A multi-pin IC chip is first being inserted to a PC board towards a nominal insertion pose in (i) and (ii). If the insertion fails (iii) as detected by force feedback, the robot lifts the chip (iv), updates goal insertion pose, and retries the insertion. We marked target hole locations in white in (iii) and (v). We define the Cartesian coordinate system with $\mathrm{X}$, $\mathrm{Y}$, and $\mathrm{Z}$ axis as shown in (i). (b) shows a sample of target components with various specifications $\epsilon$.}
\label{fig:task}
\vspace{-10pt}
\end{figure}

This paper presents the design and hardware implementation of a composable framework that tackles the above challenges. 
To handle task variations as well as unperceivable defects, our prior work \cite{CoRL2020} proposed a tolerance-guided policy learning method, which 1) parameterizes the policy using the tolerance to ensure transferrability of the policy to new components and 2) maintains a tolerance-guided probablistic model to infer the best insertion pose based on failed insertion attempts in order to ensure adaptability of the policy to defective components. 
The method has been validated in simulation but yet to be implemented on real hardware. The challenges regarding hardware limitations and collision-induced pin deformation remain unsolved, which motivate our design of the proposed composable framework. In particular, we designed an optimal configurable Cartesian tracker (OCCT) to better interface with hardware under communication delay, and a safe insertion agent (SIA) to handle the collision. The probabilistic inference module (PIM) in \cite{CoRL2020} is further extended to handle perception error. In addition, to overcome the sim2real gap~\cite{koos2012transferability} and enable safe and data efficient learning on hardware, we design a novel safe learning module (SLM) using evolutionary methods. The performance of the proposed method is evaluated on a UR10 robot.

\begin{figure}[t]
\centering
\includegraphics[width=\columnwidth]{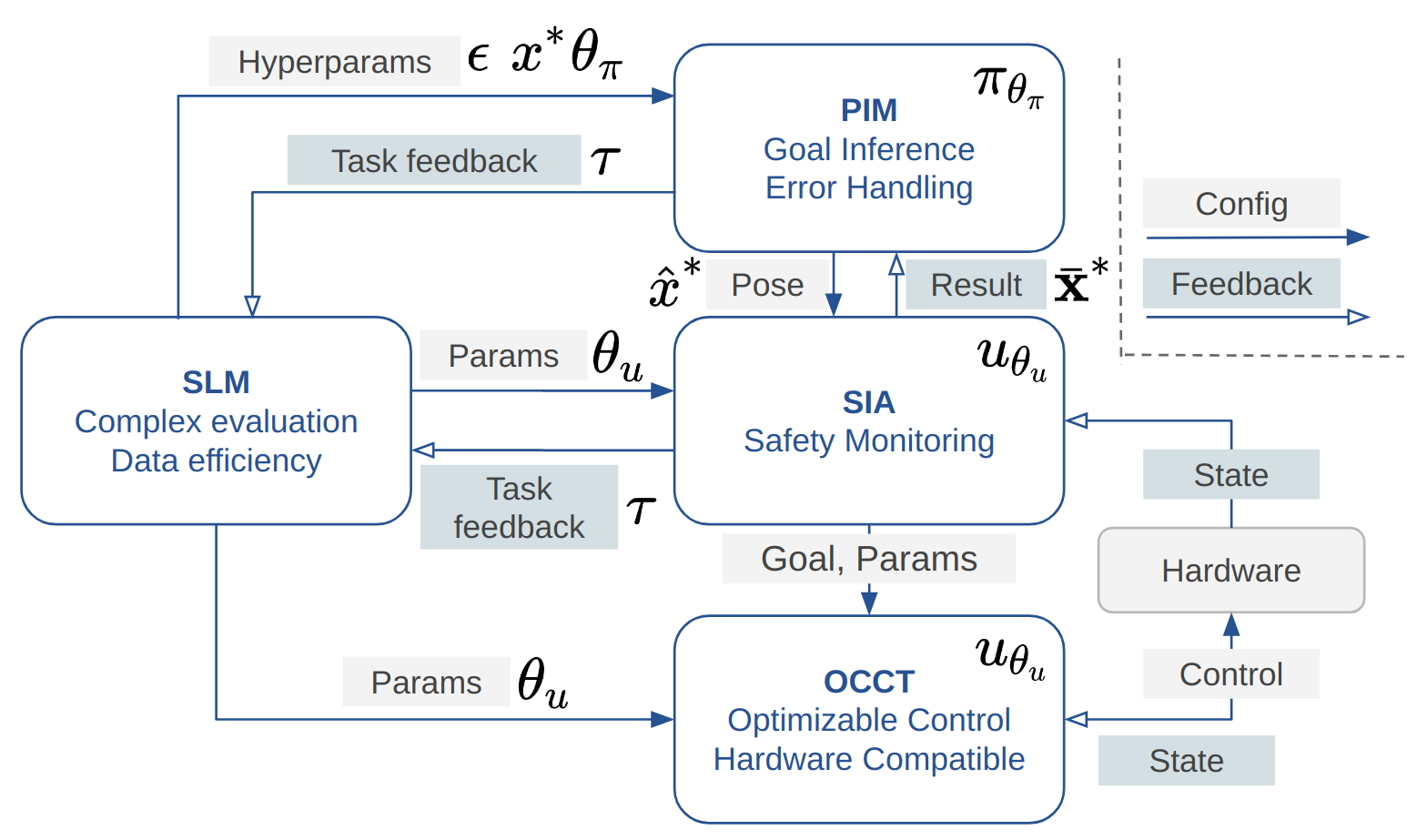}
\caption{Our insertion framework consists of four modules. Flow of module inputs and feedbacks are depicted in solid and hollow arrows respectively. Variables that are passed between modules are also marked (see \cref{sec:arch} for detailed description). The probabilistic inference module (PIM) generates insertion routines and send to the safe insertion agent (SIA). The SIA guides the insertion process by publishing Cartesian goals and handling failures. The optimal configurable Cartesian tracker (OCCT) directly communicates with the robot to perform optimization-based Cartesian space control. Finally, the safe learning module (SLM) optimizes the whole framework for task-specific objectives.}
\label{fig:framework}
\vspace{-10pt}
\end{figure}


\section{Related Work}\label{sec:related_work}


\subsubsection*{Policy learning for robotic insertion tasks}
Recently, learning-based methods have been applied to robotic insertion on hardware \cite{drl_high_prec_assemb, learning_for_high_prec_assemb, rl_var_impedance_control_prec_assemb}. 
These approaches achieve high-precision assembly, but only consider robust components with limited variations. This paper considers delicate multiple-contact insertion with variations in more aspects such as component geometry and insertion tolerance.

\subsubsection*{Safe and data efficient learning on hardware}
In this paper, we perform learning directly on hardware, where one immediate challenge is the data efficiency and safety. In literature, deep reinforcement learning-based approaches demonstrate limited data efficiency~\cite{schoettler2019deep}. 
Even though the data efficiency can be improved \cite{vecerik2019practical}, there is no guarantee on safety.
On the other hand, principled logic-based policies can guarantee safety, which are commonly used in industrial applications. Since these policies usually have much fewer parameters than deep neural networks, directly learning with these policies can ensure data efficiency as well as safety. These learning methods are usually referred to as automatic parameter tuning \cite{Chen_1988}. Nonetheless, these policies are not always differentiable, hence unable to learn with gradient-based methods. This paper introduces evolutionary algorithms \cite{Doncieux_Mouret_Bredeche_Padois_2011} to learn these parameters. 

\subsubsection*{Goal inference under uncertainties}
When there are perception errors and component defects, the actual insertion goal (as the relative position between the perceived board and the component) may deviate from the desired nominal insertion goal, and is not directly perceivable. 
To identify goal locations without visual sensors, existing methods (mostly in single peg-hold-insertion) utilize blind search, which first rub the peg across the surface to search for a partially-inserted pose \cite{park2017compliance,abdullah2015approach,jasim2014position}, then performs goal identification through pattern matching \cite{chhatpar2005particle,luo2017novel,kim2012hole}. 
These methods do not apply to delicate multi-contact insertion since the delicate components have little tolerance for shear forces during surface rubbing. 
Instead, we perform the goal inference from failed insertion attempts probabilistically \cite{CoRL2020}. 

\section{Problem Specification and System Architecture}\label{sec:arch}


\subsection{Problem Specification}

We consider delicate industrial insertion tasks where a robotic arm inserts multi-pin components into PC boards with an attached gripper as illustrated in \cref{fig:task}. The definition of the Cartesian coordinates is shown in \cref{fig:task}(a). The $\mathrm{Z}$ axis points upwards vertically while the $\mathrm{X}$ and $\mathrm{Y}$ axes are parallel to the PC board. We denote the Cartesian pose of the gripper as $x\defeq[t_x^\top, \theta_x^\top]^\top\in\mathbb{R}^6$ where $t_x\in\mathbb{R}^3$ is the translation vector and $\theta_x\in\mathbb{R}^3$ is the rotation vector in angle-axis representation. Each insertion task is encoded as a tuple $\langle\epsilon, x^*\rangle\sim \mathcal{E} \times\mathcal{X}^*$ where $\epsilon$ refers to the component specifications (e.g., geometries, insertion tolerance, etc.) and $x^*$ refers to the nominal insertion pose calculated from task specification and workspace perception. For example, each component in \cref{fig:task_variance} would correspond to a different $\epsilon$ and \cref{fig:task_overall}(ii) shows one possible nominal insertion pose $x^*$. Notably, different component instances from the same task $\epsilon$ can have undetectable variations in the physical form due to manufacturing defect. Besides, the workspace perception is prone to errors. As such, the nominal insertion pose $x^*$ might not be feasible for every component instance. This requires our insertion policy to detect whether $x^*$ is feasible for each particular component and infer an alternative insertion pose $\hat{x}^*$ if necessary. Hence, we learn two functions: (a) an adaptation policy $\pi_{\theta_{\pi}}:\bar{\mathbf{x}}^* \times x^*\mapsto \hat{x}^*$ that proposes the next insertion pose $\hat{x}^*$ from the insertion history $\bar{\mathbf{x}}^*$ (as a list of states that result in collision) and the nominal goal $x^*$ and (b) an insertion policy $u_{\theta_{u}}:\hat x^*\mapsto \tau$ that generates arm trajectories to track $\hat{x}^*$. $\theta_{\pi}$ and $\theta_{u}$ are adjustable policy parameters. Our objective is to maximize the expected reward $R$ based on the executed trajectory $\tau$ over the task distribution $\mathcal{E} \times\mathcal{X}^*$:
\begin{equation}
    \underset{\theta_\pi, \theta_u}{\mathrm{max}}~\mathbb{E}_{\langle\epsilon, x^*\rangle\sim \mathcal{E} \times\mathcal{X}^*}\Big[\mathbb{E}_{\hat{x}^*\sim\pi_{\theta_{\pi}}(\bar{\mathbf{x}}^*, x^*)}\big[\left.R\left(\tau\right)\right|_{\tau=u_{\theta_u}(\hat{x}^*)}\big]\Big]. \label{optim:prob_spec}
\end{equation}
The inner expectation in \eqref{optim:prob_spec} quantifies the insertion performance on a specific component insertion task $\langle\epsilon, x^*\rangle$ where the actual insertion pose $\hat{x}^*$ is adapted by $\pi_{\theta_\pi}$ and tracked by controller $u_{\theta_u}$. The reward function $R$ can be customized to evaluated the executed trajectory $\tau$ based on various performance indices such as insertion time, whether a collision happens, and maximum contact force on the component. Finally, the outer expectation in \eqref{optim:prob_spec} allows the insertion framework to be optimized over the task distribution $\mathcal{E} \times\mathcal{X}^*$. Specifically, $\mathcal{X}^*$ covers all possible insertions poses on the PC board (see \cref{fig:task_overall}(ii) for an example) and $\mathcal{E}$ includes all possible components (see \cref{fig:task_variance} for examples). 

There are several sub-goals we aim to achieve when solving problem \eqref{optim:prob_spec}: 1) minimizing the learning efforts for the robot by incorporating as much prior knowledge as possible in the policy design; 2) minimizing the manual integration and tuning efforts by instantiating automatic parameter tuning on hardware; and 3) accounting for limitations on resources by enabling flexible scheduling of computation resources for efficient hardware execution.
    

\subsection{Framework Architecture}\label{sub_sec:framework_arch}

The proposed composable industrial insertion framework is shown in \cref{fig:framework}. As an overview, the optimization objective in \eqref{optim:prob_spec} is evaluated together by three modules: the probabilistic inference module (PIM), the safe insertion agent (SIA), and the optimal configurable Cartesian tracker (OCCT). The PIM constitutes the adaptation policy $\pi_{\theta_\pi}$ while the SIA and OCCT together consitute insertion policy $u_{\theta_u}$. Then, the overall optimization \eqref{optim:prob_spec} is solved by a safe learning module (SLM). Next, we elaborate on each of these modules and describe how they interact to solve \eqref{optim:prob_spec}.

Given an insertion task $\langle\epsilon, x^*\rangle$, the PIM $\pi_{\theta_{\pi}}$ proposes an insertion pose $\hat{x}^*$ and passes it to the insertion policy $u_{\theta_u}$.
Inside $u_{\theta_u}$, the SIA first generates a sequence of sub-goals which starts at current gripper pose $x$ and ends with the insertion pose $\hat{x}^*$. The OCCT then controls the arm to track each sub-goal in the sequence while monitoring collision. When the component collides with the board, the SIA notifies the PIM to adjust $\hat{x}^*$ and repeats the insertion attempts. An insertion succeeds when  the $Z$ position of $\hat{x}^*$ is reached without collision and $XY$ position errors are within tolerance. An insertion fails when a maximum number of failures occur or when the component is determined as un-insertable if the maximal probability for a successful insertion drops below a threshold. One insertion loop in SIA terminates when either the insertion succeeds or it fails. 
The executed trajectory $\tau$ is then evaluated by a customizable reward function $R$. Finally, the SLM may optimize SIA, OCCT, and PIM to maximize $R$. The SLM is able to learn both offline and online. During offline training tasks, the SLM learns general configurations of these modules. During online execution, the SLM can continue to fine-tune those configurations based on newly received tasks. 

The composable framework achieves the sub-goals in the following way.
First, the framework is able to incorporate prior knowledge in SIA regarding how to perform insertion (to be elaborated in \cref{subsec:sia}). Second, the system parameters can be automatically learned using SLM (to be discussed in \cref{subsec:slm}). Third, the modules to can scheduled to run at different frequency to maximize the usage of computation resources. 
\section{Module Design}\label{sec:module}

This section provides a detailed discussion on each module in our insertion framework in terms of their functionalities and the challenges that they address. We also discuss the transferability of our framework to other manipulation tasks in terms of efforts required.

\subsection{Safe Insertion Agent}\label{subsec:sia}

\begin{figure}[t]
\centering
\includegraphics[width=\columnwidth]{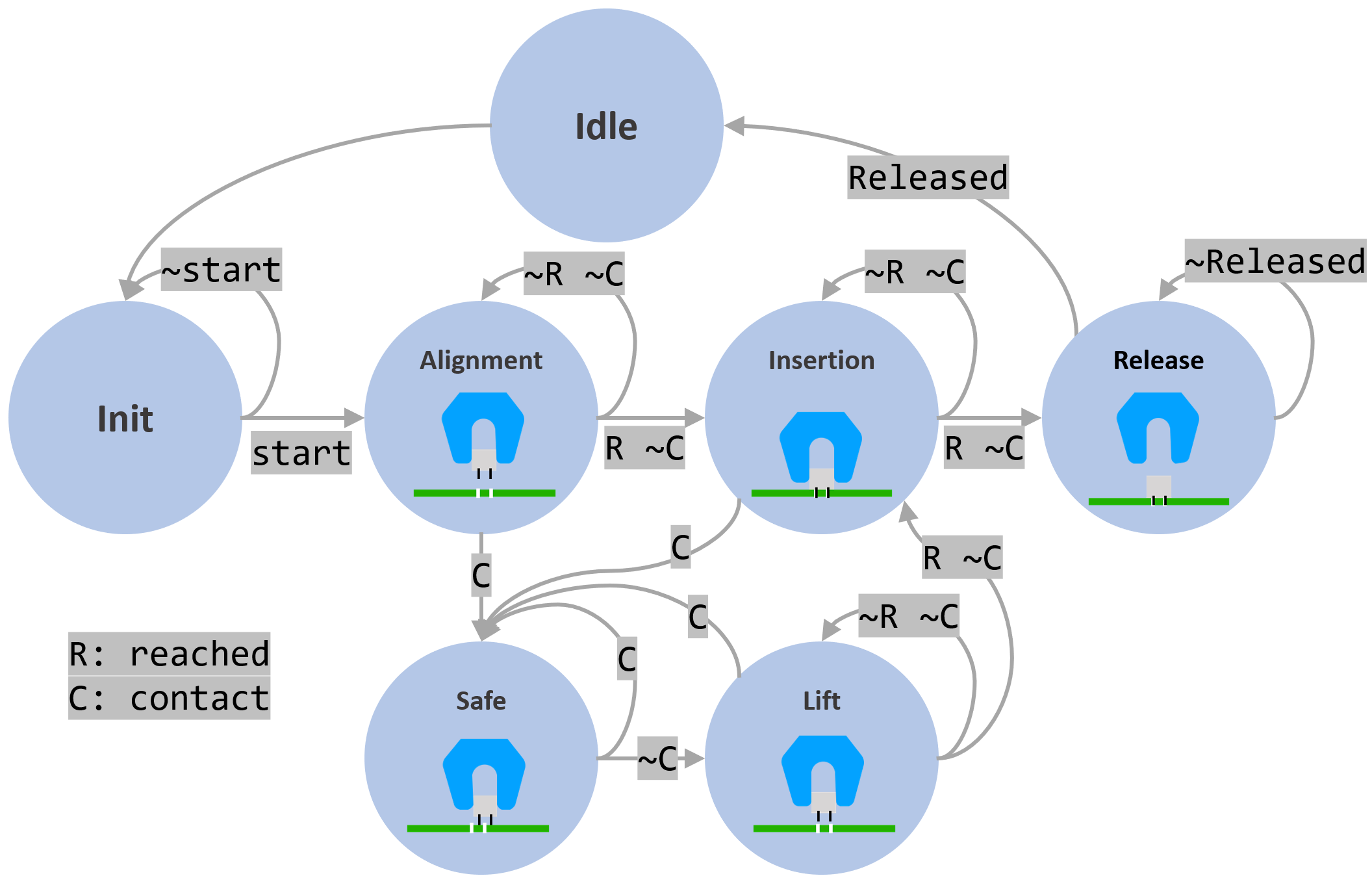}
\vspace{-15pt}
\caption{The SIA initially stays in \textit{init} state and wait for triggering signal. When triggered, the SIA sequentially commands the alignment pose and insertion pose. When the pins collide with the board, the SIA enters \textit{safe} state and lifts the component. After receiving updated goals from PIM, SIA re-enters \textit{insertion} state to make the next attempt. Once the insertion succeeds, the SIA releases the component and resets to the \textit{init} state. The whole logic-based policy is invoked per SIA control loop where the state transits according to the criteria marked on the links.}
\label{fig:sia_state_machine}
\vspace{-15pt}
\end{figure}

We design the safe insertion agent (SIA) to generate and execute insertion routines and handles component-board collisions. In the insertion task shown in \cref{fig:task}, it is helpful to apply a logic-based policy for robustness and intepretability. One rule we summarize from observing human demonstrations is that in order to insert a multi-pin chip, one needs to first align the pins above the holes and then move downwards while keeping all the pins perpendicular to the board. Another rule is that when the pins make contact with the board, we lift the component to prevent damage and minimize the shear force. We incorporate those two rules as a logic-based policy which guides the insertion routine in SIA (see \cref{fig:sia_state_machine} for the state machine). When a state transition is triggered either by collision or by fulfillment of goal reaching, the SIA sends the next tracking reference to the OCCT.

The logic-based policy in SIA requires an external specification of two key poses: the \textit{alignment pose} $x^a$ and the \textit{insertion pose} $\hat{x}^*$. $x^a$ differs from $\hat{x}^*$ only in the $Z$ component, i.e., $x^a=\hat{x}^*+\Delta z\ \vec Z$ where $\Delta z$ is the \textit{alignment offset}. We assume that the component is already grasped by the gripper when the insertion task starts. The SIA first enters \textit{alignment} state and commands $x^a$. Then, the SIA enters \textit{insertion} state and commands insertion pose $\hat{x}^*$. If any pin-board collision is detected during the movements, implying that $x^a$ and $\hat{x}^*$ need corrections, the SIA enters \textit{safe} state and lifts the component. The collision is reported to an external module (the PIM in our case) which then corrects $x^a$ and $\hat{x}^*$. Then, the SIA returns to \textit{insertion} state and re-tries the insertion with updated goals. This loop terminates under the conditions mentioned in \cref{sub_sec:framework_arch}. After that, the SIA releases the gripper, broadcasts the execution statistics, and reset to the \textit{init} state.

Our SIA design enables motion-wise behavior configuration. The alignment state closes the major gap between initial position and target position with negligible chance of collision with the PC board, thus can be executed in high speed with generous tolerance on tracking error. The insertion state performs final insertion, thus should be optimized for stable and accurate control in lower speed. Such flexibility helps the framework to achieve both safety and efficiency.

\subsection{Cartesian Tracking Control for Robot Arms}

The optimal configurable Cartesian tracker (OCCT) controls the robot arm via the Robot Operating System (ROS) control interface \cite{ros_control} to track a Cartesian reference commanded by the SIA. We design OCCT as an  optimization-based controller to achieve two major objectives: (a) robustness to hardware communication delays and (b) optimality of the generated trajectories. We desire the robot to track references in Cartesian space via least possible effort (e.g., torques) within time limits. In this work, we use a UR10 robot and such optimality is not guaranteed by current drivers such as UR ROS driver\footnote{\label{fn:UR10}https://github.com/UniversalRobots/Universal\_Robots\_ROS\_Driver} and Moveit\footnote{https://github.com/ros-planning/moveit}.



Communication delay exists in any ROS-based system. The delay is intractable and unpredictable since it depends on real-time network conditions. Thus, it is difficult to implement a reactive OCCT control loop with high sampling frequency. 
Nonetheless, a $125\mathrm{Hz}^{\ref{fn:UR10}}$ control loop is required for smooth arm motion. This motivates us to isolate OCCT and the arm driver. Instead of doing reactive control, we plan and send a whole trajectory that goes to the target pose specified by SIA. The trajectory is then executed by the arm driver$^{\ref{fn:UR10}}$ with high sampling frequency. To maintain responsiveness to insertion events as discussed in \cref{subsec:sia}, OCCT refreshes at $10\mathrm{Hz}$, the same rate as SIA. However, it will only modify an existing trajectory when the target pose specified by SIA changes or when there is a large tracking error. Note that the second case never occurs since the tracking error is always small to be shown in the experiments in \cref{sec:ctc_exp}.

To achieve trajectory optimality, we first generate the Cartesian trajectory by solving a ${{T}}$-step discrete linear-quadratic regulator (LQR) problem and then solve inverse kinematics (IK) for the corresponding joint trajectory. In the LQR problem, the Cartesian state $x_k\defeq[t_{x_k}^\top, \theta_{x_k}^\top]^\top$ evolves according to a second-order dynamics under control input $u_k=\ddot{x}_k$ with time interval $\Delta t$. The objective is to drive $x_k$ to the reference which is assumed to be the origin for simplicity. We define the run-time cost as $l_k(x_k, u_k)=\frac{1}{2}[x_k^\top, \dot{x}_k^\top]Q[x_k^\top, \dot{x}_k^\top]^\top+\frac{1}{2}u_k^\top Ru_k$ and the  terminal cost as $ l_N(x_{{T}})=\frac{1}{2}[x_{{T}}^\top, \dot{x}_{{T}}^\top]S[x_{{T}}^\top, \dot{x}_{{T}}^\top]^\top$ where $Q=v_Q\cdot\mathbf{I}$, $S=v_S\cdot\mathbf{I}$, and $R=\mathbf{I}$. $\mathbf{I}$ refers to identify matrices. There is an input constraint $u_k\in[-b_u, b_u]$ (to bound acceleration), an orientation constraint $\theta_{x_k}=\mathbf{0}$ (to ensure that the end-effector points downward vertically), a velocity constraint $\dot{x}_k\in[-b_v,b_v]$, and a terminal state constraint $x_{{T}}=\mathbf{0}$. The corresponding optimization problem is defined as
\begin{align}
    \underset{u_{0:{{T}}-1}}{\mathbf{min}} &~\sum_{k=0}^{{{T}}-1}l_k(x_k, u_k) + l_{{T}}(x_{{T}}) \label{eq: mpc_regulation}\\
    \mathbf{s.t.} &~\begin{bmatrix}x_{k+1}\\\dot{x}_{k+1}\end{bmatrix} = \begin{bmatrix} \mathbf{I} & {\Delta t}\cdot\mathbf{I} \\ 0 & \mathbf{I} \end{bmatrix} \begin{bmatrix}x_{k}\\\dot{x}_{k}\end{bmatrix} + \begin{bmatrix} 0.5{\Delta t}^2\cdot\mathbf{I} \\ {\Delta t}\cdot \mathbf{I} \end{bmatrix} u_k \nonumber\\
    &~u_k\in[-b_u, b_u],~\dot{x}_k\in[-b_v,b_v],~\theta_{x_k}=\mathbf{0},~x_{{T}}=\mathbf{0} \nonumber
\end{align}
Solving \eqref{eq: mpc_regulation} yields the optimal planned trajectory $\mathbf{x}\defeq\{x_0,\dot{x}_0,x_1,\dot{x}_1,\cdots,x_{{T}},\dot{x}_{{T}}\}$.
Denote the robot joint state as $q\in[-\pi,\pi]^{|q|}$ where ${|q|}$ is the number of joints. We then solve IK to convert $\mathbf{x}$ to joint space as $\mathbf{q}\defeq\{q_0,\dot{q}_0,q_1,\dot{q}_1,\cdots,q_{{T}},\dot{q}_{{T}}\}$ using the ICOP algorithm~\cite{zhao2020contactrich}. Notably, the IK is solved for Cartesian poses within a short range from the current state. As such, we can avoid singularity of IK by choosing insertion poses far from singularities. The OCCT finally sends the joint trajectory command $\mathbf{q}$ to the UR10 ROS driver for execution.

All OCCT parameters can be optimized by SLM to achieve user-defined goals. For example, the choices of planning horizon ${{T}}$ and time interval $\Delta t$ depends on the computation resources and scale of the task environment. A larger ${{T}}$ improves reachability of the trajectory at the cost of longer planning time that grows exponentially, while a smaller $\Delta t$ improves trajectory smoothness at the cost of reduced reachability. 
Since an optimal OCCT configuration can hardly be chosen manually, it is sensible to learn them using SLM.

\subsection{Probabilistic Inference Module}
The insertion goal has uncertainty due to possible component defection and perception errors. In our previous work \cite{CoRL2020}, we proposed a model to handle component defection. The model infers the adapted insertion goal $\hat{x}^* = \pi_{\theta_{\pi}}(\bar{\mathbf{x}}^*, x^*)$ that the model believes to have the best chance to succeed based on the insertion history $\bar{\mathbf{x}}^*$ and the nominal goal $x^*$. In that case, we assume no perception error and the perception $x^*_t$ exactly matches $x^*$, the true insertion goal with which a defect-free component can be inserted. 
Our prior model solves the following optimization:
\begin{equation}
    \underset{\hat{x}^*}{\mathrm{max}}~\mathbb{E}_{\langle\epsilon,x^*\rangle\sim\langle\mathcal{E}|_{x^*},x^*\rangle}\Big[\mathbf{1}(\epsilon,\hat{x}^*)\mid \bar{\mathbf{x}}^*\Big],    \label{optim:pm_old}
\end{equation}
where $\mathcal{E}|_{x^*}$ represents the distribution of possible defects for that particular component which is known in the manufacturing process and $\mathbf{1}(\epsilon,\hat{x}^*)$ indicates whether the adapted insertion goal $\hat{x}^*$ is feasible. The above optimization constitutes the mapping $\pi_{\theta_{\pi}}:\bar{\mathbf{x}}^* \times x^* \mapsto \hat{x}^*$.

This paper considers perception error which can cause the perceived nominal goal $x^*_t$ to deviate from the true nominal goal $x^*$. Thus, we extend the probabilistic inference to incorporate perception-induced goal uncertainty. It is assumed that the true nominal goal follows a Gaussian distribution given the perceived goal: $x^*\sim\mathcal{N}(x^*_t, \Sigma)$ where the covariance $\Sigma$ approximates the perception uncertainty and is tunable by SLM. This paper only considers the variance in $\mathrm{X}$ and $\mathrm{Y}$ directions denoted as $\Sigma_x$ and $\Sigma_y$. Given the perceptual uncertainty, we need to construct a new policy $\pi_{\theta_{\pi}}:\bar{\mathbf{x}}^* \times x^*_t \mapsto \hat{x}^*$ that solves the following optimization:
\begin{equation}
    \underset{\hat{x}^*}{\mathrm{max}}~\mathbb{E}_{\langle\epsilon,x^*\rangle\sim\langle\mathcal{E}|_{x^*},x^*\rangle,x^*\sim\mathcal{N}(x^*_t,\Sigma)}\Big[\mathbf{1}(\epsilon,\hat{x}^*)\mid \bar{\mathbf{x}}^*\Big].
    \label{optim:pm_new}
\end{equation}
We calculate the expectation in \eqref{optim:pm_new} via sampling. The sample size $N_{\text{smp}}$ influences both the estimation accuracy and the computational effort. Large sample size results in more accurate estimation but also takes longer time to compute. Similar trade-off also applies to the optimization process. Since we utilize CMA-ES \cite{hansen2016cma} to optimize for the best $\hat{x}^*$, the generation size $N_{\text{gen}}$, population size $N_{\text{pop}}$, and elite size $N_{\text{elite}}$ all influence the accuracy of the outcome $\hat{x}^*$ and the computational effort. 
In summary, $\theta_\pi$ contains the following parameters: $N_{\text{smp}}$, $N_{\text{gen}}$, $N_{\text{pop}}$, and $N_{\text{elite}}$, $\Sigma_x$ and $\Sigma_y$. These parameters can all be learned by SLM instead of selected manually. We will empirically show that PIM leads to less insertion attempts until success than random search in \cref{sec:sia_pm_exp} and leave formal analysis of success rate as future work.

\subsection{Safe Learning Module}\label{subsec:slm}
The safe learning module (SLM) is designed to optimize the safety and efficiency of the insertion framework. The framework has many deeply coupled parameters that are difficult for hand tuning. Moreover, these parameters are not directly differentiable which precludes gradient-based learning. We choose CMA-ES as our learning algorithm since it is derivative free \cite{hansen2016cma}. Briefly speaking, to solve \eqref{optim:prob_spec}, CMA-ES iteratively samples a set of candidate parameters $\theta_\pi$ and $\theta_u$ and keep only those who have a high \textit{fitness} score based on the evaluation of the objective in \eqref{optim:prob_spec}. During training, we keep the environment configuration (e.g., robot, camera, etc.) the same since the parameters learned by CMA-ES would be invalid if the environment changes. We leave learning under changing environments as future work.

To further improve the efficiency of learning, we introduce the following two approaches: decomposition of the parameter space and augmentation of hardware learning with learning in simulation. Due to the curse of dimensionality, the learning complexity and the learning time grow exponentially with the number of parameters. Therefore, to improve the efficiency of learning, we divide the parameters to learn into groups where strongly coupled parameters are in the same group. Our composable framework design naturally suggests the division of parameters into two groups: one containing parameters of OCCT and SIA and the other containing parameters of PIM. The two groups are decoupled since the first group fully controls the execution of insertions (as $u_{\theta_u}$ in \eqref{optim:prob_spec}) while the second only controls goal generation (as $\pi_{\theta_\pi}$ in \eqref{optim:prob_spec}).


Besides parameter decomposition, we also augment hardware learning with learning in simulation to improve data efficiency. The motivation is that, it is impractical to train part of the modules directly on hardware due to intrinsic needs for large amounts of data. For example, learning PIM parameters requires insertions with diverse defective components. In addition, switching between components is time consuming and requires massive human labor. To mitigate this issue, we learn PIM parameters in simulation first, and then fine-tune the learned parameters on hardware. Notably, transferring policies learned in simulation to hardware is difficult due to the {sim2real gap}, i.e., the simulation cannot accurately model the real-world physics. With our composable framework, however, the sim2real gap is mitigated since we can train each module separately and use different learning platforms. Specifically, we perform simulation learning only on the PIM since it does not require high-fidelity physics simulation. On the other hand, we train OCCT and SIA directly on hardware to avoid the sim2real gap. 


\subsection{Transfer to Other Manipulation Tasks}

The composable design of our framework enables transfer to other manipulation cases with appropriate modifications. Specifically, both SIA and PIM need to be re-designed to accommodate different manipulation requirements. The methods presented above is specific to multi-pin component insertion. On the other hand, both OCCT and SLM are task-agnostic can be directly applied to other manipulation tasks. OCCT provides a general robot control interface that facilitates policy execution on various hardware, while SLM is able to optimize the policy in different environments as long as an objective is available like in \eqref{optim:prob_spec}.

\section{Experiments}\label{sec:exp}

This section presents hardware experiments with IC sockets on a UR10 robotic arm to demonstrate how our composable design satisfies user specifications under various constraints. 
The reason why we chose IC sockets is that the pins of IC chips are extremely delicate and are bent ``out of the box''. The remedy that human operators have to take is to align and flatten them before inserting them. We slightly simplified the problem by focusing on IC sockets, which are more robust but can still bend under large contact force. Since the shape of the IC socket body is very similar to that of the chip, it is not an oversimplification of the insertion task.
In our experiments, we use a Ensenso X36 stereo camera kit to perceive the PC board and components to insert. The camera kit generates high-resolution point clouds that lead to localization standard deviations within $0.1~\mathrm{mm}$. Moreover, we record actual robot trajectories from the UR10 ROS driver which we consider as the ground truth.

Section \ref{sec:ctc_exp} shows how OCCT accommodates limited hardware control API and achieves high precision motion execution. Section \ref{sec:sia_pm_exp} demonstrates how SIA and PIM interact to handle perception errors and component defects during insertion, which also validate the safety of the system in handling collisions. Section \ref{sec:slm_exp} shows that SLM can efficiently optimize other modules and provide insights on how different parameters are coupled and how they affect the insertion performance. The adaptability and transferrability of the method to new tasks is not directly shown unlike in \cite{CoRL2020}, but can be supported by the short amount of time it needs to fine-tune the system on hardware. 


\subsection{Tracking in Cartesian Space}\label{sec:ctc_exp}

\begin{figure}[t]
    \begin{center}
        \resizebox{\columnwidth}{!}{\input{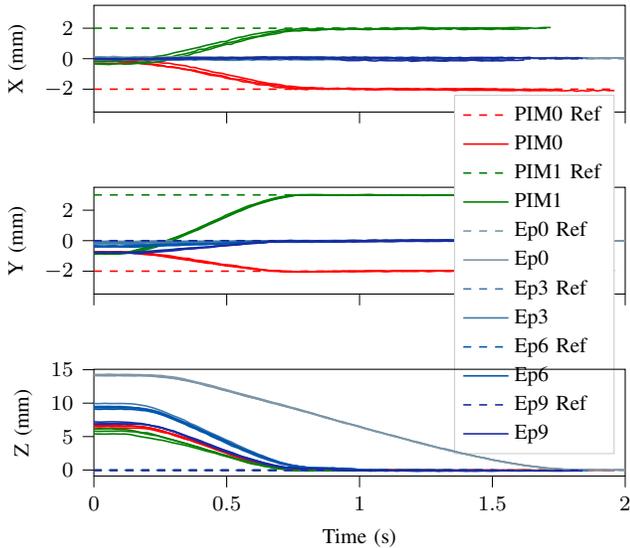}}
        \vspace{-20pt}
        \caption{Gripper trajectories in $\mathrm{X}$, $\mathrm{Y}$ and $\mathrm{Z}$ positions during four training insertions (from different SLM learning epoches), i.e., Ep0, Ep3, Ep6, and Ep9, and two PIM-guided insertions, i.e., PIM0 and PIM1. The dotted lines are the target positions set by SIA, while the solid curves are the executed trajectories. Three repeated insertions are conducted for each task and are plotted in the same color to show repeatability. In training tasks, all goal poses share the same $\mathrm{X}$ and $\mathrm{Y}$ component. In PIM-guided tasks, desired $\mathrm{X}\mathrm{Y}$ positions are changed by PIM due to goal position inference.}
        \label{fig:ctc_exp}
    \end{center}
    \vspace{-20pt}
\end{figure}

This section shows the static reference tracking of OCCT under several settings from two scenarios. First, we demonstrate that OCCT can generate optimal trajectories during the learning process where the OCCT parameters change. Then, we demonstrate that OCCT can generate optimal trajectories under different goals $\hat{x}^*$  proposed by the PIM. When solving the Cartesian trajectory $\mathbf{x}$, we applied constraints $\ddot{x}_{max}=b_u=0.02~\mathrm{m}/\mathrm{s^2}$ and $\dot{x}_{max} = b_v = 0.02~\mathrm{m}/\mathrm{s}$.
All trajectories are shown in \cref{fig:ctc_exp}. We repeated each experiment three times to demonstrate repeatability of the trajectories. During learning, the OCCT parameters and insertion heights $\Delta z$ were changed in each epoch by SLM. We can see that in all epochs (from grey to blue curves), the OCCT generated and tracked smooth trajectories while maintaining linear motions (maximum $10^{-4}~\mathrm{m}$ horizontal error along $\mathrm{X}$ and $\mathrm{Y}$).
Thus, OCCT was able to solve optimal trajectories while guaranteeing safety constraints. Besides, when guided by PIM, the OCCT was also able to track $\mathrm{X}\mathrm{Y}$ references smoothly (red and green curves). In all experiments, OCCT showed repeatable trajectories, meaning that the OCCT well separated the generation of task trajectories from actual robot hardware execution.


\begin{figure}[t]
    \includegraphics[width=\columnwidth]{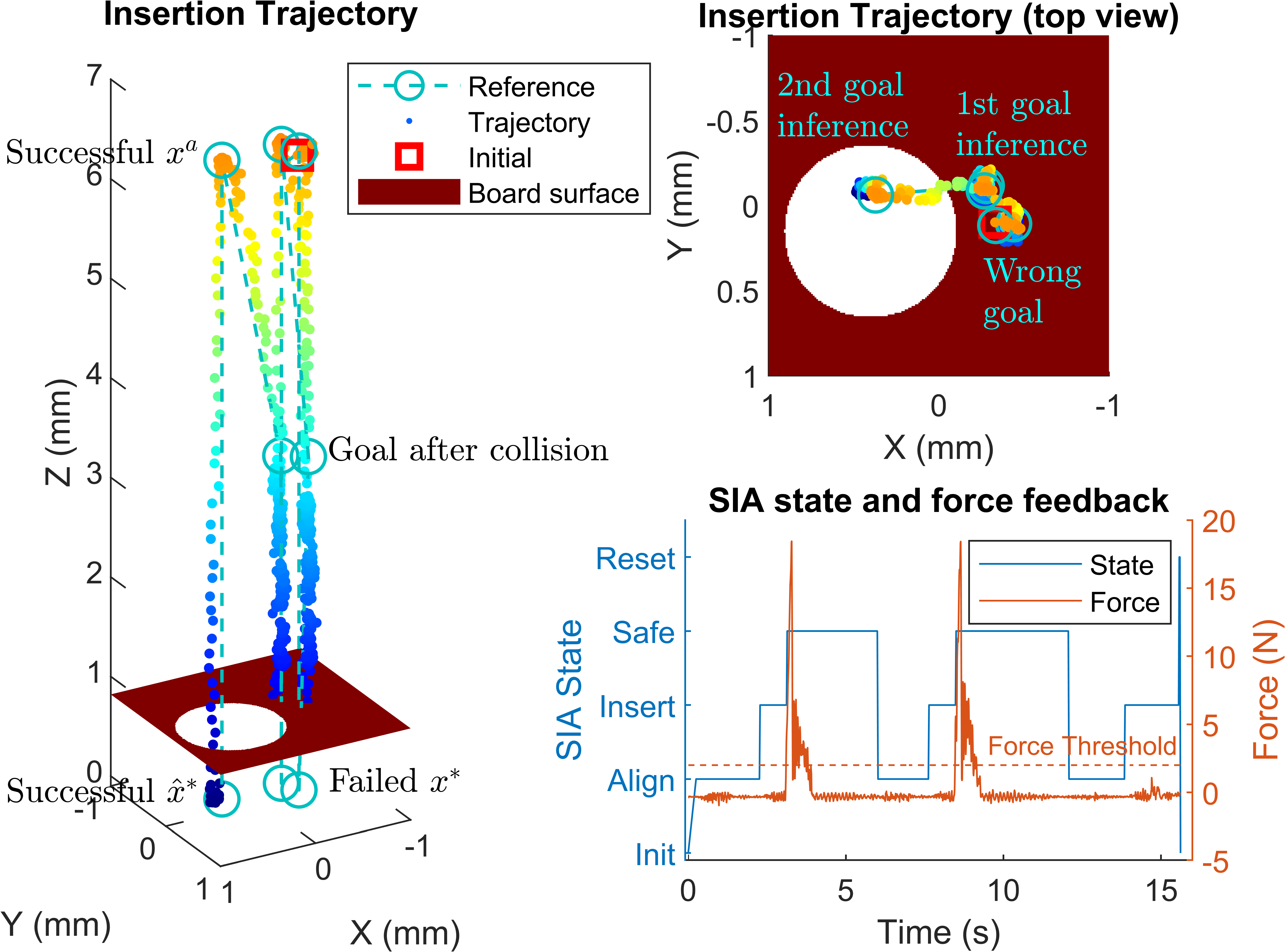}
    \caption{Insertion with with inaccurate nominal insertion pose $x^*$. The left figure shows the gripper trajectory in configuration space. A top-down view is also provided. When $x^*$ was inaccurate, the component collided with the board where the contact force exceeded the threshold $2~\mathrm{N}$ (bottom right). The SIA entered safe state and lifted the gripper by commanding intermediate goals above the collision position. After PIM updated $\hat{x}^*$ (see top right for how the inferred goal evolved in $\mathrm{X}$-$\mathrm{Y}$ plane), the SIA made the next attempt. After two goal inference steps, the insertion succeeded. 
    }
    \label{fig:pm_sia}
    \vspace{-10pt}
\end{figure}

\subsection{Single Component Insertion with Multiple Attempts} \label{sec:sia_pm_exp}

This section demonstrates how SIA and PIM guided the insertion of an IC socket. We used an IC socket with $2\times 4$ pins (circle pins with $0.6~\mathrm{mm}$) and applied our framework to insert it into a PC board with $1.0~\mathrm{mm}$ diameter holes. Other types of sockets are also applicable since PIM can adapt to different components as described in \cite{CoRL2020}. To demonstrate error handling, we arbitrarily chose an inaccurate nominal insertion pose ${x}^*$ (see \cref{fig:pm_sia}) and bent one of the socket pins slightly. This simulated how perception errors and component defection fail insertions. During the insertion, we monitored the contact force on the pins via a force torque sensor mounted on the gripper. A collision would be alarmed when the contact force exceeds $2~\mathrm{N}$. This threshold could be adjusted depending on the robustness of the pins and the board. When collision happened, SIA entered the \textit{safe} state, lifted the socket, and notified the PIM. The PIM then inferred the next insertion goal based on failed insertions and updates $\hat{x}^*$ according to \eqref{optim:pm_old}. After receiving the update, the SIA made the next insertion attempt. \Cref{fig:pm_sia} illustrates the gripper trajectory and the SIA states throughout the process. In that example, after two collisions, the PIM located the real insertion position within insertion tolerance and the insertion succeeded. We see that even with erroneous goal specifications, SIA and PIM could detect insertion failures and make proper adjustments to accommodate those errors. Notably, the maximum instantaneous pin-board contact force was less than $20~\mathrm{N}$ and any force larger than $5~\mathrm{N}$ lasts only for very short amount of time. This was within safety range since we did not observe collision-induced pin deformation.
In addition, we compared PIM to a random policy which sampled new insertion locations from a Gaussian distribution centered at $x^*$. Results showed that PIM took $2.8$ insertions on average with an advantage of $0.9$ over the random policy.
\subsection{Automatic Parameter Tuning} \label{sec:slm_exp}

In this experiment, we used SLM to optimize the parameters of OCCT, SIA, and PIM to improve the insertion efficiency and safety. Benefit from our composable design, the parameters to learn (see \cref{tab:SLM exp}) can be divided into a low-level group and a high-level group. Each group can be learned separately. The low-level set consists of OCCT control parameters and insertion distance. The high-level set consists of PIM parameters which controls the goal inference procedure. To ease the learning process on hardware, we first learned each parameter set separately and then fine-tuned them jointly. Next, we describe the SLM optimization objectives and analyze the learning results to answer the following questions: (a) how much can SLM improve the performance, (b) how are the module parameters coupled or decoupled and how do they affect the insertion efficiency and safety, and (c) how data-efficient and scalable is SLM.

\begin{table}[t]
    \centering
    \begin{tabular}{c l c c  c}
    \toprule
         Module & Parameters & Default & Range & Result\\
         \midrule
         OCCT & $v_Q$ & 10 & [0, 20] & 13.102\\
         OCCT & $v_S$ & 100 & [90, 110] & 91.605\\
         OCCT & $T_{\text{OCCT}}$  & 1 & [0.75, 3] & 0.75\\
         OCCT & $1/\Delta t$ & 10 & [5, 20] & 9.278\\
         SIA & $\Delta z$ & 0.015 & [0.005, 0.025] & 0.0056\\
         \midrule
         PIM & $N_{\text{smp}}$ & 500 & [5, 1000] & 334\\
         PIM & $N_{\text{gen}}$ & 10 & [2, 20] & 2\\
         PIM & $N_{\text{pop}}$ & 50 & [3, 100] & 44\\
         PIM & $N_{\text{elite}}$ & 15 & [3, 30] & 15\\
         PIM & $\Sigma_x$ & 0.0003 & [0.00001, 0.0005] & 0.000405\\
         PIM & $\Sigma_y$ & 0.0003 & [0.00001, 0.0005] & 0.000363\\
         \bottomrule
    \end{tabular}
    \caption{Parameters to learn.}
    \label{tab:SLM exp}
    \vspace{-20pt}
\end{table}

\subsubsection{OCCT \& SIA Learning on Hardware} \label{sec:ctc_learning}

\begin{figure}[t]
    \begin{center}
    \begin{tabular}{c}
        \resizebox{\columnwidth}{!}{
\begin{tikzpicture}

\definecolor{color0}{rgb}{0.75,0.897651006711409,1}
\definecolor{color1}{rgb}{0.12156862745098,0.466666666666667,0.705882352941177}
\definecolor{color2}{rgb}{1,0.867219917012448,0.75}
\definecolor{color3}{rgb}{1,0.498039215686275,0.0549019607843138}

\pgfplotsset{
    layers/my layer set/.define layer set={
        background,
        main,
        foreground
    }{
    },
    set layers=my layer set,
}

\begin{axis}[
height = 30mm,
width = \linewidth,
tick align=inside,
tick pos=left,
every axis title/.style={at={(0.5,1.1)}},
title = (a) OCCT \& SIA learning,
font = \footnotesize,
x grid style={white!69.0196078431373!black},
xmin=0, xmax=250,
xtick style={color=black},
y grid style={white!69.0196078431373!black},
ylabel={Force},
ymin=-0.07750395, ymax=0.8,
ytick style={color=black}
]

\addplot [very thick, color1, on layer=foreground]
table {%
0 0.318237588235294
50 0.415135352941177
100 0.273154352941176
150 0.195596725490196
200 0.07797856
250 0.07797856
}; \label{force_line}
\addplot [draw=color0, fill=color0, mark=*, only marks, opacity=0.3]
table{%
x  y
0 0.027753
1 0.374723
2 0.383254
3 0.17496
4 0.062182
5 0.27866
6 0.339226
7 0.126791
8 0.130406
9 0.0429
10 1.176003
11 0
12 0.548727
13 0.389242
14 0
15 0.35269
16 0.431892
17 0.155387
18 0.954794
19 0
20 0.567878
21 0.481634
22 0.139233
23 0.418113
24 0
25 0
26 0.382597
27 0.46323
28 0.482761
29 0.527289
30 0.939282
31 0.408924
32 0.587949
33 0.36918
34 0.362537
35 0.369721
36 0.008434
37 0.286565
38 0.683087
39 0.150955
40 0.084394
41 0.277499
42 0.210713
43 0.304052
44 0.37454
45 0.356423
46 0.25667
47 0.239192
48 0.149896
49 0.170756
50 0.227023
51 0.257827
52 0.313463
53 0
54 0.237459
55 0.301202
56 0.240534
57 0
58 0.324561
59 0.733365
60 0.172848
61 0
62 0.293993
63 0.968912
64 0.651582
65 0.437955
66 0.476173
67 0.370535
68 0.489576
69 0.687709
70 0.468975
71 0.377706
72 0.646325
73 0.272672
74 0.496332
75 0.319779
76 0.176304
77 0.241821
78 0.69875
79 0.509524
80 0.477911
81 0.644993
82 0
83 0.549199
84 0.540613
85 0.699412
86 0.184333
87 0
88 1.078809
89 0.188961
90 0.111552
91 0.514283
92 0.488662
93 0.227964
94 0.517674
95 0.43066
96 0
97 0.96099
98 0.299133
99 0.735285
100 1.128564
101 1.008214
102 0.618717
103 0.533351
104 0.275534
105 1.06291
106 0.161116
107 0.248785
108 0.789701
109 0.747115
110 0.300201
111 0.531906
112 0.466572
113 0.476856
114 0.25761
115 0
116 0.445285
117 0.692803
118 0.352515
119 0.493202
120 0.414784
121 0.317455
122 0
123 0.298901
124 0.212478
125 0.140037
126 0.546139
127 0.051861
128 0.076534
129 0.313312
130 0.104267
131 0
132 0.26417
133 0
134 0.036883
135 0
136 0.081669
137 0.111049
138 0
139 0
140 0.17742
141 0.164808
142 0.016993
143 0
144 0.004359
145 0.006796
146 0
147 0
148 0
149 0
150 0
151 0
152 0
153 0
154 0.04544
155 0.023541
156 0
157 0
158 0
159 0
160 0
161 0.205896
162 0.013324
163 0.0322
164 0
165 0.04095
166 0.177636
167 0.032483
168 0.296646
169 0
170 0
171 0.042258
172 0
173 0.627741
174 0.090229
175 0
176 0.425697
177 0.539721
178 0.110528
179 0
180 0.363747
181 0.02222
182 0.052302
183 1.550079
184 0.24496
185 0.161764
186 0.283189
187 1.228414
188 0.250533
189 0.222351
190 0.072434
191 1.350503
192 0.476339
193 0.101139
194 0.133072
195 0.013373
196 0.12763
197 0.166595
198 0.080904
199 0.301933
200 0.067662
201 0.025931
202 0.081796
203 0.03887
204 0.123528
205 0.088296
206 0
207 0.117529
208 0.049262
209 0
210 0.008826
211 0.095069
212 0.006051
213 0.027906
214 0
215 0.168493
216 0
217 0
218 0
219 0.050759
220 0.080183
221 0.170126
222 0.06837
223 0.11938
224 0.651315
225 0
226 0
227 0
228 0.025752
229 0.26957
230 0.077397
231 0.01605
232 0.006935
233 0.036227
234 0.032481
235 0.308095
236 0
237 0.588725
238 0.006737
239 0
240 0.106044
241 0
242 0
243 0
244 0
245 0.106791
246 0
247 0.044331
248 0
249 0.234441
}; \label{force_scatter}

\end{axis}

\begin{axis}[
axis y line=right,
tick align=inside,
x grid style={white!69.0196078431373!black},
font=\footnotesize,
height = 30mm,
width = \linewidth,
xmin=0, xmax=250,
xtick pos=left,
xtick style={color=black},
y grid style={white!69.0196078431373!black},
ylabel={Time},
ymin=1.9, ymax=2.6,
ytick pos=right,
ytick style={color=black},
yticklabel style={anchor=west}
]
\addlegendimage{/pgfplots/refstyle=force_line}\addlegendentry{Average force}
\addplot [very thick, color3, on layer=foreground]
table {%
0 2.41935355135
50 2.26059432086364
100 2.16870694204255
150 2.16239244770588
200 2.17199757684
250 2.17199757684
}; \label{time_line}
\addlegendentry{Average time}

\addplot [draw=color2, fill=color2, mark=*, only marks, opacity=0.3]
table{%
x  y
0 2.81923426
1 2.744030201
2 2.599892354
3 2.319718442
5 2.167881236
6 3.815749307
7 2.073965587
8 1.976019887
9 2.53990137
10 2.039537586
12 2.28065853
13 2.006455224
15 2.31324861
16 2.198753535
17 2.454634441
18 3.459878003
20 2.471237416
21 2.539874882
23 2.239939397
26 2.639712781
27 2.583902462
28 2.262186887
29 1.919918028
30 2.879313325
31 2.530257014
32 2.287534266
33 2.287166983
34 2.339899146
35 2.515864666
37 2.01991226
38 2.658197726
39 2.600474267
41 2.239917043
42 2.13875662
43 2.339184896
45 2.161068111
46 2.307177929
47 2.384198589
48 2.719028634
49 1.899860153
51 1.879902058
52 1.950796663
54 2.199897652
55 1.999926472
56 1.979897014
58 2.435851645
59 2.079905659
60 2.079905532
62 2.759890842
63 1.899893349
64 1.919914496
65 2.679891807
66 1.953262791
67 2.443323617
68 2.103621967
69 2.405632373
70 1.999344936
71 2.219713072
72 2.623540909
73 2.194787297
74 2.679867658
75 2.0232736
76 2.007005138
77 2.039895594
78 2.399885226
79 2.535620339
80 2.219920033
81 2.539920284
83 2.699895128
84 2.439684605
85 2.67989224
86 2.213342032
88 2.139930789
89 2.038927676
90 1.66230427
91 2.316987833
92 2.179903928
93 2.080312838
94 2.262721147
95 2.279893201
97 2.099863401
98 2.659635864
99 2.798709135
100 2.659858008
101 2.479907215
102 2.519895609
103 2.659814884
104 1.839618976
105 2.185620416
106 2.059898126
107 1.704907941
108 2.859931489
109 2.879917842
110 1.6806486
111 2.065218229
112 2.259894112
113 2.159876013
114 1.939932213
116 2.267928119
117 2.176436037
118 2.361160746
119 2.263643165
120 2.359944673
121 2.104524104
123 2.819340749
124 2.160054385
125 1.942686445
126 1.839960866
127 1.917394276
128 1.918365012
129 1.97986622
130 1.921722988
131 2.135646189
132 1.999910961
133 2.256950003
134 2.072639971
135 2.139865058
136 2.007504727
137 2.099912765
138 1.932436953
140 1.959862456
141 2.121577549
142 2.095704065
143 2.059906913
145 2.33244037
146 2.112351627
147 1.870647185
148 2.299906175
149 2.15298708
150 2.291008771
151 1.839886628
152 2.32775635
153 2.468480054
154 2.199888686
155 2.079346773
156 2.119858982
157 1.887109659
158 2.279155069
159 2.17697305
160 1.942766963
161 1.942196329
162 1.959849279
163 2.049809573
164 1.868980752
165 1.979891021
166 1.998126756
167 2.039891067
168 2.174689329
169 2.079941395
170 1.859916488
171 2.036396927
172 2.45416019
173 2.559924471
174 2.15990644
175 2.139933632
176 2.459895593
177 2.551347573
178 2.719881106
179 2.019906361
180 2.538759976
181 2.01572686
182 1.62102983
183 2.199903512
184 2.159918678
185 2.514833707
186 2.059912653
187 2.135870073
188 2.150643535
189 1.866537937
190 2.045010498
191 2.388418886
192 2.255798603
193 2.380336332
194 2.189282351
195 2.121059971
196 2.336527998
197 2.039904443
198 2.1533235
199 2.152646242
200 2.289693981
201 3.159895858
202 2.339888779
203 1.959906638
204 2.251764328
205 2.423572327
206 2.079917389
207 2.319896925
208 2.239878244
209 2.397534569
210 2.320480031
211 2.519910093
212 2.239315876
213 2.319897376
214 2.268706807
215 2.059878699
216 2.200663891
217 2.639912613
218 2.054849337
219 1.599875192
220 1.87986364
221 2.239930743
222 1.959863636
223 1.999883367
224 2.451228431
225 2.058138333
226 1.728321378
227 2.09235198
228 1.999981931
229 1.879886228
230 2.591991299
231 2.059901585
232 2.013509935
233 1.765095681
234 2.363516745
235 2.176206696
236 2.039911504
237 2.145703674
238 2.370206389
239 2.040435036
240 2.202333714
241 1.967486617
242 2.339863661
243 2.041863941
244 1.883273301
245 2.215214231
246 1.74419824
247 1.924175889
248 2.11586494
249 2.624237144
};\label{time_scatter}


\end{axis}

\end{tikzpicture}} \\
        \resizebox{\columnwidth}{!}{\input{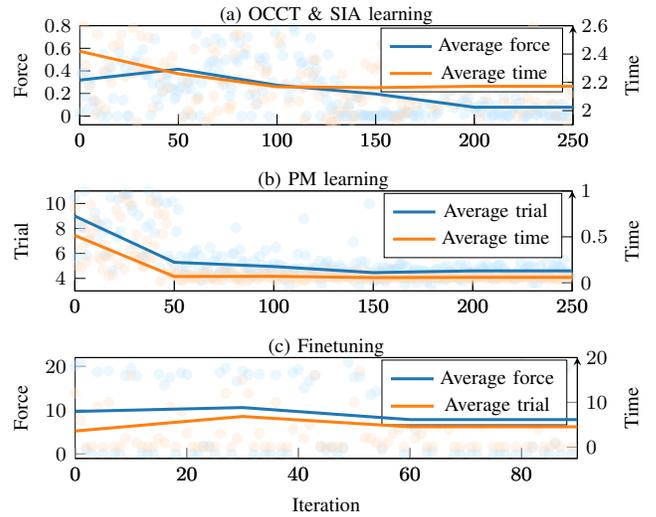}} \\
        \resizebox{\columnwidth}{!}{
\begin{tikzpicture}

\definecolor{color0}{rgb}{0.75,0.897651006711409,1}
\definecolor{color1}{rgb}{0.12156862745098,0.466666666666667,0.705882352941177}
\definecolor{color2}{rgb}{1,0.867219917012448,0.75}
\definecolor{color3}{rgb}{1,0.498039215686275,0.0549019607843138}

\pgfplotsset{
    layers/my layer set/.define layer set={
        background,
        main,
        foreground
    }{
    },
    set layers=my layer set,
}

\begin{axis}[
tick align=inside,
tick pos=left,
x grid style={white!69.0196078431373!black},
font = \footnotesize,
title= (c) Finetuning,
every axis title/.style={at={(0.5,1.1)}},
xlabel={Iteration},
height=30mm,
width=\linewidth,
xmin=0, xmax=90,
y grid style={white!69.0196078431373!black},
ylabel={Force},
ymin=-1.0470154, ymax=21.9873234,
]

\addplot [very thick, color1, on layer=foreground]
table {%
0 9.684848
30 10.5779285806452
60 7.82484743333333
90 7.82484743333333
}; \label{force_line}
\addplot [draw=color0, fill=color0, mark=*, only marks, opacity=0.3]
table{%
x  y
0 20.058414
1 20.769461
2 0.932234
3 0
4 18.6968
5 0.105509
6 18.958749
7 0.13055
8 0
9 17.86587
10 19.457731
11 17.795009
12 0.311313
13 0
14 0
15 18.202308
16 15.780858
17 17.458871
18 0
19 18.508649
20 18.077326
21 0
22 0
23 1.767252
24 1.74335
25 0
26 18.56893
27 19.556415
28 0
29 16.583948
30 18.900741
31 18.15858
32 0
33 18.183308
34 0
35 0
36 18.006669
37 0.229025
38 0
39 18.522138
40 19.232919
41 15.061253
42 16.631553
43 17.932017
44 1.815666
45 0.795291
46 1.606212
47 0
48 1.242626
49 20.940308
50 17.503784
51 17.657182
52 7.46432
53 13.565282
54 14.295044
55 0.925629
56 14.557606
57 16.857141
58 18.295751
59 0
60 19.535741
61 5.484579
62 16.44855
63 0.421134
64 0.544487
65 17.76675
66 18.943238
67 19.34158
68 0
69 0
70 18.496157
71 0
72 0
73 17.605973
74 19.615092
75 0
76 0
77 0
78 17.106769
79 0
80 20.531367
81 0
82 1.597764
83 18.608708
84 0.526507
85 0
86 2.272325
87 0.030166
88 19.682151
89 0.186385
}; \label{force_scatter}

\end{axis}

\begin{axis}[
axis y line=right,
tick align=inside,
x grid style={white!69.0196078431373!black},
font=\footnotesize,
height=30mm,
width=\linewidth,
xmin=0, xmax=90,
xtick pos=left,
y grid style={white!69.0196078431373!black},
ylabel={Time},
ymin=-2.55, ymax=20,
ytick pos=right,
yticklabel style={anchor=west}
]
\addlegendimage{/pgfplots/refstyle=force_line}\addlegendentry{Average force}
\addplot [very thick, color3, on layer=foreground]
table {%
0 3.58064516129032
30 6.83870967741935
60 4.53333333333333
90 4.53333333333333
}; \label{time_line}
\addlegendentry{Average trial}

\addplot [draw=color2, fill=color2, mark=*, only marks, opacity=0.3]
table{%
x  y
0 12
1 6
2 0
3 0
4 9
5 0
6 10
7 0
8 0
9 2
10 16
11 1
12 0
13 0
14 0
15 2
16 1
17 9
18 0
19 8
20 3
21 0
22 0
23 0
24 0
25 0
26 3
27 12
28 0
29 2
30 15
31 12
32 0
33 3
34 0
35 0
36 10
37 0
38 0
39 4
40 49
41 2
42 5
43 16
44 0
45 0
46 0
47 0
48 0
49 51
50 10
51 17
52 2
53 2
54 1
55 0
56 1
57 3
58 4
59 0
60 5
61 2
62 3
63 0
64 0
65 5
66 10
67 4
68 0
69 0
70 11
71 0
72 0
73 5
74 28
75 0
76 0
77 0
78 15
79 0
80 15
81 0
82 0
83 4
84 0
85 0
86 1
87 0
88 28
89 0
};\label{time_scatter}


\end{axis}

\end{tikzpicture}}
    \end{tabular}
    \end{center}
    \vspace{-10pt}
    \caption{Figures from top to bottom show curves for: OCCT \& SIA learning, PIM learning, and fine-tuning. Every scatter point corresponds to a set of parameters. Infeasible data points are not counted in the average. In OCCT \& SIA learning and PIM learning, the metrics all reduced over iterations, indicates that the safety and efficiency were improved. In fine-tuning, the metrics did not change significantly, suggesting that the parameters from different groups are well decoupled. Therefore joint fine-tuning did not improve the performance. We did not consider perception error in OCCT \& SIA learning, so there was no collision and the forces were caused by robot movements. However, larger forces in fine-tuning were caused by collisions.}
    \label{fig:curves}
\end{figure}

We designed the following reward to jointly optimize the parameters: planned execution time $T_{\text{OCCT}}={T}\cdot\Delta t$, planning frequency $1/\Delta t$, LQR cost $v_Q$, $v_S$, and alignment offset $\Delta z$:
\begin{align}
    R = - t_{\text{insert}}^2 - 10\ C - 0.1\ F - 100\ \mathbf{1}_\text{infeas} - 100\ \mathbf{1}_\text{fail}, \label{eq:ctc_reward}
\end{align}
where $t_{\text{insert}}$ is the insertion time, $C$ is the collision count, $F$ is the maximal contact force, $\mathbf{1}_\text{infeas}$ indicates failure of planning, and $\mathbf{1}_\text{fail}$ indicates failure of insertion. $t_{\text{insert}}$ is the total time SIA spends in the \text{insertion} state for reaching a single $\hat{x}^*$ which includes the computation time of OCCT trajectory, execution time of the trajectory on hardware, and communication delays. The goal is to reduce the cycle time as much as possible while keeping the insertion safe and feasible. 
The learning results are shown in \cref{fig:curves} (a), which shows that both the force and the time decrease with iterations. As for the parameters shown in \cref{tab:SLM exp}, $v_Q$ is increased, while $v_S$, $T_{\text{OCCT}}$, $1/\Delta t$, $\Delta z$ are decreased.

\begin{figure}[t]
    \begin{center}
        
        \begin{tabular}{cc}
            \includegraphics[width=0.45\linewidth]{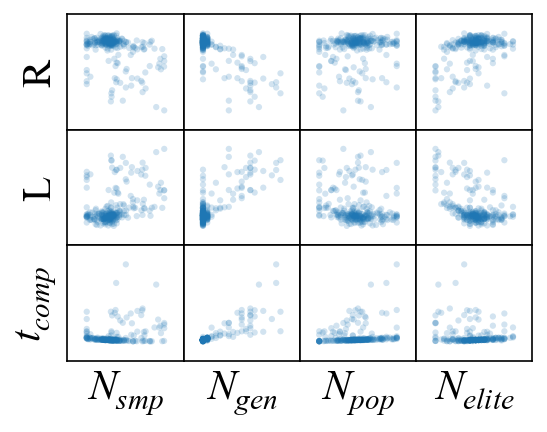} &
            \includegraphics[width=0.5\linewidth]{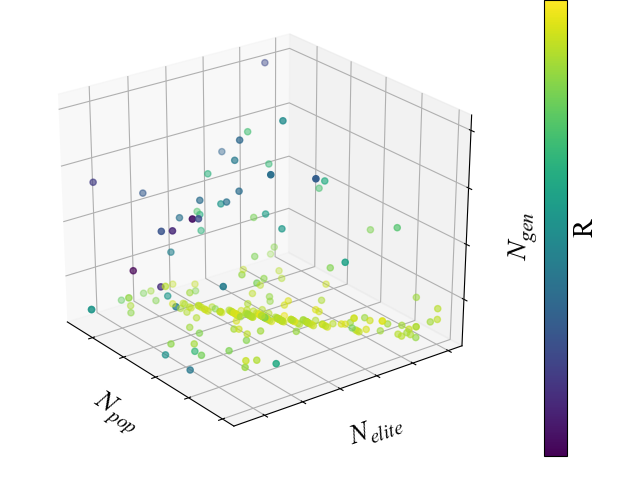} \label{fig:high_dim}
        \end{tabular}
            
        
        \vspace{-5pt}
        \caption{Illustration of parameter coupling. Left: (a) Correlation analysis between parameters and metrics in PIM learning. Parameters are listed on the X axis, while metrics are listed on the Y axis. Each scatter point corresponds to one experiment. Most parameters do not show strong correlations to metrics. 
        Right: (b) Reward distribution with regard to $N_{\text{gen}}$, $N_{\text{pop}}$, and $N_{\text{elite}}$. The reward shows a strong correlation to the elite-population ratio.}
            \label{fig:SLM_corr}
    \end{center}
    \vspace{-15pt}
\end{figure}

\subsubsection{PIM Learning in Simulation}
The parameter $\theta_\pi$ in PIM needs to be learned to balance the accuracy of the inferred insertion pose and its computation time. It worth noting that we have a good initial estimation of the uncertainty $\Sigma_x$ and $\Sigma_y$, which are directly obtained from hardware specifications. So we skipped them in this phase to reduce the learning time, and left them for the fine-tune step. 
We tested the PIM parameters on 100 randomly generated defected chip insertion poses. If PIM infers the correct insertion pose within 20 trials, we consider the insertion successful. The reward function is defined as:
\begin{align}
    R = -100\ \mathbf{1}_\text{fail} - 2\ L - t_{\text{comp}},
\end{align}
where $t_{\text{comp}}$ is the computation time of PIM, and $L$ is the number of insertion trials. 
The learning results are shown in \cref{fig:curves} (b), which shows that the average insertion trial and average computation time were both reduced. As for the parameters  shown in \cref{tab:SLM exp}, $N_{\text{gen}}$ was greatly reduced by the learning. This suggests a low $N_{\text{gen}}$ is enough for accurate pose estimation. $N_{\text{smp}}$ and $N_{\text{pop}}$ were slightly decreased to reduce $t_{\text{comp}}$, and $N_{\text{elite}}$ remained unchanged.

\subsubsection{Fine-tuning on Hardware}

We jointly fine-tuned all the parameters to verify our sim-to-real transfer and examine undiscovered coupling between different parameter groups. If \textit{the reality gap} exists, the parameters learned in simulation can be further improved by fine-tuning on real hardware. Similarly, if there is undiscovered coupling between parameters from different groups, the performance will also be improved. To increase fine-tuning efficiency, the exploration range is reduced to around $20\%$ of the range in learning since the parameters are already near optimal. 
We used the same reward function defined in \cref{eq:ctc_reward}, because this reward reflects the whole insertion task performance. As shown in \cref{fig:curves} (c), the performance does not show a significant change, which indicates that the parameters learned in simulation were successfully transferred to reality, and the parameters were well decoupled across groups.
Notably, the force readings in OCCT \& SIA learning phase was smaller than those during finetuning. This is because in the first case, we only provided feasible $x^*$ to ensure collision-free insertions. As such, the force readings came from acceleration of the arm. In the finetuning phase, we deliberately fed infeasible $x^*$ to cause collisions, resulting in contact forces in the $10~\mathrm{N}$ to $20~\mathrm{N}$ range.

\subsubsection{Parameter analysis}
We inspected the coupling among parameters by analyzing their correlations to different metrics in $R$. From \cref{fig:SLM_corr} (a), we can see that the correlations between individual parameters and metrics are not obvious. But there are strong correlations when we visualize their joint distribution as shown in \cref{fig:SLM_corr} (b). This suggests that the coupling exists and parameter tuning must be conducted jointly. Some parameters show a strong correlation to certain metrics, such as $N_{gen}$ and $t_{comp}$. Hence, we can learn $N_{gen}$ separately from others to further reduce learning time. 


\begin{table}[]
    \centering
    \begin{tabular}{c c}
    \toprule
       Exp.  &  Time\\
    \midrule
       OCCT \& SLA learning (hardware) & ~30 min\\
       PIM learning (simulation) & ~180 min\\
       Fine tuning (fine-tune) & ~60 min\\
    \bottomrule
    \end{tabular}
    \caption{Learning time.}
    \label{tab:learning_time}
    \vspace{-20pt}
\end{table}

\subsubsection{Learning efficiency}
The learning time is shown in \cref{tab:learning_time}. PIM learning was performed on a PC with AMD Ryzen threadripper 3960x 24-core processor. No parallel computing was used. But PIM learning can be easily sped up by parallel computing. Other experiments were performed on a UR10 robot with a PC host equipped Intel Xeon(R) E5-2609 v4 1.7GHz 8 core CPU.

\section{Conclusion}\label{sec:conclusion}

This paper presented a composable framework for safe and efficient industrial insertion. The framework features four individual modules, each addressing different challenges: the OCCT for optimal insertion trajectory generation under constraints, the SIA for collision handling, the PIM for goal inference under component defection and perception errors, and the SLM for learning on hardware. We applied our framework on a delicate socket insertion task with a UR10 robotic arm. We showed that the framework was able to generate and execute optimal insertion trajectories under safety constraints, handle failed insertions and infer true insertion positions, and automatically tune module parameters for safety and efficiency on hardware. For future work, we will theoretically analyze the robustness of the proposed framework.





\section*{ACKNOWLEDGMENT}
This work is supported by Efort Intelligent Equipment Co., Ltd. The authors would like to thank Brad Lisien, Dan Troniak, Yueqi Li, Weiye Zhao, Charles Noren, and Ruixuan Liu for their help in experiments, and Joseph Giampapa for his help in polishing the paper.

\bibliographystyle{IEEEtran}
\bibliography{IEEE}

\end{document}